\documentclass[runningheads]{llncs}

\pdfoutput=1
\usepackage{graphicx}
\usepackage{amsmath,amssymb} % define this before the line numbering.
\usepackage{color}
\usepackage{footnote}
\usepackage[width=122mm,left=12mm,paperwidth=146mm,height=193mm,top=12mm,paperheight=217mm]{geometry}

\newcommand{\mynote}[1]{\renewcommand{\thefootnote}{\alph{footnote}}\footnotemark[#1]}

\begin{document}
% \linenumbers
\pagestyle{headings}
\mainmatter
\title{Noise generation for compression algorithms} % Replace with your title
%
%\titlerunning{ECCV-18 submission ID \ECCV18SubNumber}
%
%\authorrunning{ECCV-18 submission ID \ECCV18SubNumber}
%
%\author{Anonymous ECCV submission}
%\institute{Paper ID \ECCV18SubNumber}

\titlerunning{Noise generation for compression algorithms}
\authorrunning{Renata Khasanova, Jan Wassenberg, Jyrki Alakuijala}

\author{Renata $\text{Khasanova}^\text{a}$, \quad Jan $\text{Wassenberg}^\text{b}$, \quad Jyrki $\text{Alakuijala}^\text{b}$\\
	{\tt\small renata.khasanova@epfl.ch, \quad \{janwas, jyrki\}@google.com}}

\institute{\mynote{1}{\hspace{1pt} LTS4, \'{E}cole Polytechnique F\'{e}d\'{e}rale de Lausanne} \\ \mynote{2}{\hspace{1pt} Google Research}}

\maketitle

\begin{abstract}

In various Computer Vision and Signal Processing applications, noise is
typically perceived as a drawback of the image capturing system that ought to be
removed. We, on the other hand, claim that image noise, just as texture, is important for
visual perception and, therefore, critical for lossy compression algorithms that tend to make
decompressed images look less realistic by removing small image details.
In this paper we propose a physically and biologically inspired
technique that learns a noise model at the encoding step of the compression
algorithm and then generates the appropriate amount of additive noise at the
decoding step. Our method can significantly increase the realism of the
decompressed image at the cost of few bytes of additional memory space
regardless of the original image size. The implementation of our method is open-sourced
and available at \url{https://github.com/google/pik}.
%The link to the open-sourced code for the paper
%will be added.

\end{abstract}


%-------------------------------------------------------------------------
\section{Introduction}
\label{sec:intro}

\begin{figure}[t!]
\begin{tabular}{cccc}
\includegraphics[height=3.15cm]{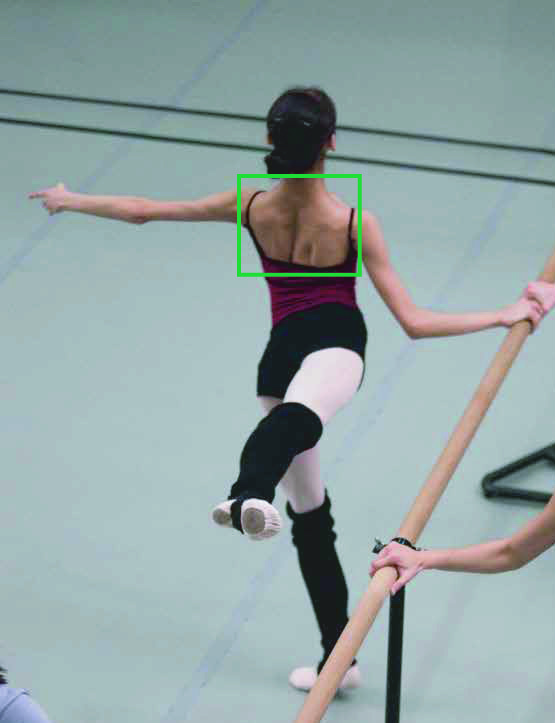} &
\includegraphics[height=3.15cm]{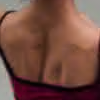}&
\includegraphics[height=3.15cm]{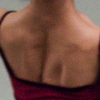}&
\includegraphics[height=3.15cm]{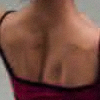} \\
(a)&(b)&(c) & (d)
\end{tabular}
\caption{Compression algorithms often remove details which are crucial for
perception. Here, (a) is an original image from which we extract the patch
marked by the green rectangle, (b) is the image patch filtered with a compression
algorithm PIK, (c) is the  original image patch, (d) is the compressed
image patch with the proposed additive noise. (best seen in color)}
\label{fig:teaser}
\end{figure}

Lossy compression techniques~\cite{bb:wallace1992jpeg,bb:haffner1999djvu,bb:pik}
allow a significant reduction of image sizes, which is highly beneficial for
various applications that involve storage~\cite{bb:dropbox,bb:g_cloud} and/or
sharing image content over the internet~\cite{bb:g_photos}. This, however, is
achieved at the cost of removal of some image details which can be clearly seen
in Fig.~\ref{fig:teaser}, where the left image illustrates the result of the
compression-decompression process of  the recently
introduced PIK~\cite{bb:pik} algorithm, applied to the middle image of the
figure. Some details have disappeared and the image looks
unnaturally smooth. To overcome this problem we suggest augmenting the
compressed image with noise that will make the resulting image more visually
appealing, as can be seen in the right image~of~Fig.~\ref{fig:teaser}.

The most straightforward way of doing this is to simply generate white noise
and add it to the decompressed image. This, however, typically results in even
less naturally looking images. %, as can be seen in
%Fig.~\ref{fig:ig:simplenoise}.
One of the main reasons is that different parts of an image
captured by an ordinary camera exhibit different amount of noise. For natural
raw images, the noise has typically smaller magnitude in the presence of a low
intensity signal and higher otherwise. Most modern cameras, however,
perform gamma-correction after capturing the image, therefore the noise level
becomes higher for a low-intensity signal and smaller otherwise.
Fig.~\ref{fig:teaser} illustrates this
phenomenon. To address this issue, we propose to use physically inspired
intensity dependent noise model \cite{bb:mandel1959fluctuations}, which
describes the interaction between the light and the camera sensor. This allows
us to model the amount of noise that we add to the compressed image as a
function of image intensities in a small neighbourhood of each pixel.

Further, our approach operates in receptor color space
\cite{bb:opsin}, which models different processes happening in the human
eye and is also used in the recent PIK \cite{bb:pik} compression algorithm. This
allows us to generate realistically looking colored noise that is
consistent with image content.

In short, in this paper, our contribution is two-fold:
\begin{itemize}
	\item Our physically and biologically inspired noise generation algorithm
  significantly improves realism of decompressed images.
	\item Our approach is reasonably fast in the decompression step, which is crucial for compression
  algorithms.
\end{itemize}
% TODO(renatakh) add more interesting table of contents
%The paper is organized as follows. In Section \ref{s:rw} we discuss
%related work. Further, we describe our algorithm in Section \ref{s:alg}.
%Finally, we demonstrate our results in Section \ref{s:exp} and conclude in
%Section \ref{s:concl}.

\section{Related work}
\label{s:rw}

To the best of our knowledge, the problem of noise re-generation for compression
algorithms is relatively unexplored.
Therefore, in this section we discuss
methods for noise estimation and film-grain effect generation tasks, as we find
them the most similar to our problem.

\paragraph{Noise estimation.} Image de-noising, lossy image compression and
other image processing algorithms~\cite{bb:liu2006noise,bb:wallace1992jpeg}
often rely on noise estimation methods as part of their pipeline. Many of these
methods assume that the amount of noise is independent for each pixel in the
image and follows the normal distribution (i.e. white gaussian noise). For
example, in order to estimate the appropriate level of image noise that needs to
be added to the image, \cite{bb:donoho1995noising} uses mean absolute deviation
and the authors in \cite{bb:tai2008fast} propose the method based on Laplacian and
Sobel filters. Contrary to these methods, we design a model that depends on the pixel
intensities, and mimics the physical and biological processes happening inside
both the human visual system and a camera sensor.

A similar direction was taken by the authors of
\cite{bb:liu2006noise,bb:liu2014signal}. The first work designs an
intensity-dependent noise model. Their method infers the noise level from a single
image using Bayesian MAP inference, which is relatively slow and therefore
cannot be directly applied to our problem.
Further, \cite{bb:liu2014signal} introduces a segmentation-based algorithm,
which also requires heavy computation to find clusters.
By contrast, we propose a fast method for
estimation of the signal-dependent noise model. Further, to make this model
biologically and physically plausible, we suggest working in XYB color
space (that is thoroughly discussed in Section \ref{s:alg_xyb}).

Additionally, the authors of \cite{bb:gardenas2017denoising} propose a noise
re-generation algorithm for video compression. This work is similar in spirit to ours,
however, the authors apply additive white
Gaussian noise, which may introduce image artifacts. Instead, we propose to process
the random signal with a high-pass filter, which allows generating a more appealing
noise for visual perception.

\paragraph{Film-grain.}
Being clearly seen in traditional analog movies, film-grain noise is currently
used as an artistic effect that makes compressed images/videos more visually
pleasing \cite{bb:yan1997signal,bb:sun2014dct} and appear as if they were
recorded on photographic film. Though the idea of making compressed images
visually pleasing is similar to our approach, the nature of film-grain noise
is very different with respect to the one coming from a digital sensor (that
we simulate), which in turn results in differently looking images. In this
section we nevertheless describe some of the approaches that allow generating
additive film-grain noise.
As such, the authors in \cite{bb:yan1997signal} generate signal-dependent
film-grain noise using higher-order statistics of the image signal.
While effective, their approach relies on noisy measurement of the
high-order signal statistics and works in RGB color space, which makes it
difficult to combine different levels of image noise coming from different RGB
channels. A different approach was proposed by \cite{bb:sun2014dct}, where the
authors use spectral domain analysis to generate film-grain noise. Their method, however,
relies on the DCT transform \cite{bb:ahmed1974discrete} for noise model
estimation, which is a time-consuming procedure and requires a large amount of
memory space to store all the DCT coefficients.

To sum up, various noise estimation models have been proposed as part of
de-noising and image compression algorithms. These models, however, either
assume the independence of noise distribution across the image or rely on
complex inference models, which are impractical for various applications.
Furthermore, there is a separate class of methods aimed at generating film-grain noise, which
however, does not aim at simulating sensor artifacts that are present in modern
digital images. Contrary to all these methods, we propose a fast and memory efficient algorithm for
noise level estimation and re-generation that  mimics the image artifacts, which
appear due to the nature of the sensors of digital cameras.

\section{Algorithm}
\label{s:alg}

\begin{figure}[t!]
\centering
\includegraphics[width=10cm]{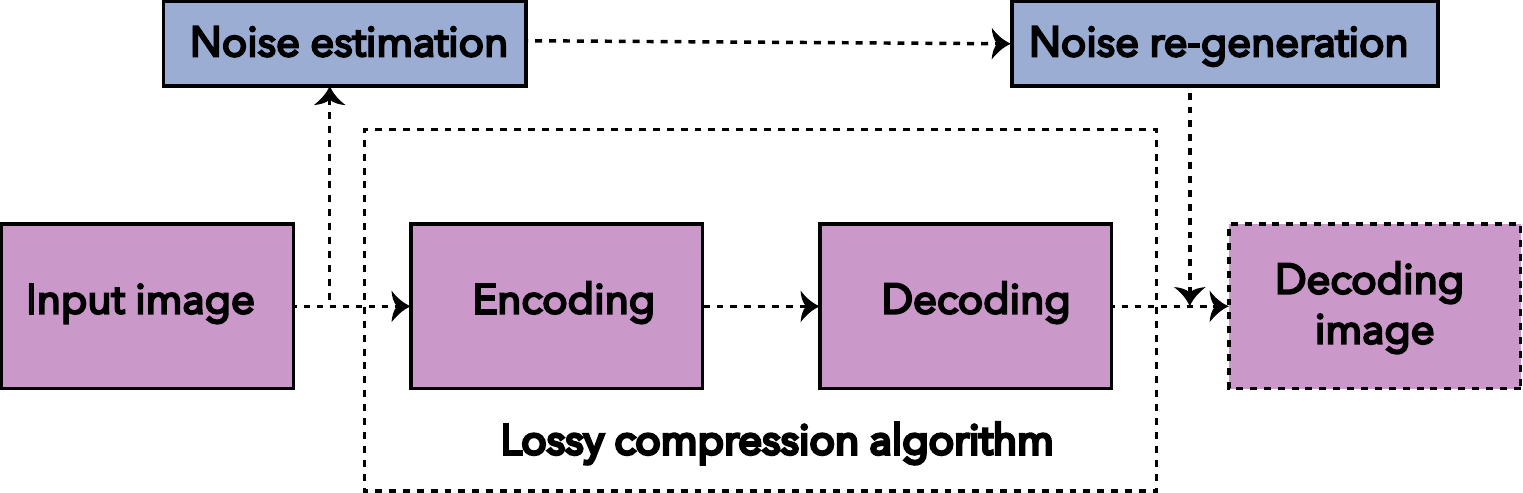}
\caption{Overview of the proposed image noise re-generation for compression
algorithm. Our method consists of two main parts -- first, we learn parameters
of the noise model, then we use this model to re-generate noise and add it to the decompressed image}
\label{fig:overview}
\end{figure}

In this section we introduce our noise re-generation algorithm. The overview of
our approach is depicted by Fig.~\ref{fig:overview}. Briefly, the algorithm
consists of the two following parts. First, prior to image encoding, we estimate
the parameters of our noise model based on the non-overlapping image patches.
We then use this model to re-generate an appropriate amount of
noise for different parts of the decompressed image.

In this section, we first introduce the color space in which our method operates and then discuss
in more detail each of the aforementioned steps.

\subsection{XYB color space}
\label{s:alg_xyb}

The human eye relies on two types of cells, rods and cones, that capture
light coming from the environment. Rods are very sensitive and capture the
intensity of the signal, while cones extract chromatic information. Cones
are themselves subdivided into three different types, which roughly capture
signals of long, medium and short wavelengths. The recently introduced XYB color
space \cite{bb:opsin} is specifically designed to model this behavior. Its core advantage
for us with respect to the commonly used color spaces RGB~\cite{bb:RGB} and
CIELAB~\cite{bb:CIELAB} is that it allows to model noise the same way as it appears in the human eye,
which in turn allows adding naturally looking augmentation to the decompressed images,
which makes them more visually pleasing.

More formally we can define the relationship between XYB and RGB as follows.
First we divide the input linear-light RGB signal into three different ones that capture
long~($L$), medium~($M$) and short~($S$) wavelengths \cite{bb:LMS} as follows:
\begin{equation}
\begin{bmatrix}
I_L \\ I_M \\ I_S\\
\end{bmatrix}
=
  \frac{1}{255}
  \begin{bmatrix}
      0.355 & 0.589 & 0.056 \\
      0.251 & 0.715 & 0.034 \\
      0.092 & 0.165 & 0.743 \\
  \end{bmatrix}
  \begin{bmatrix}
  I_R \\ I_G \\ I_B\\
  \end{bmatrix}
  ,
\end{equation}
\noindent
where $I_L, I_M, I_S$ linearly depend on photon counts \cite{bb:photons} for each cell in camera sensor
and $I_R, I_G, I_B \in [0, 255]$ are the intensities in RGB color space. Then to model the processes
happening in the human eye we apply gamma correction:
\begin{equation}
\begin{bmatrix}
I_L' \\ I_M' \\ I_S' \\
\end{bmatrix}
=
  \begin{bmatrix}
  {I_L}^\frac{1}{3} \\ {I_M}^\frac{1}{3} \\ {I_S}^\frac{1}{3}\\
  \end{bmatrix}
  .
\label{eq:lms_xyb}
\end{equation}
\noindent
For simplicity we refer to the space of these three signals (channels) as the
g-LMS space. Finally, we calculate XYB image channels as follows:
\begin{equation}
\begin{split}
I_X &= 0.5 (H_L I_L' - H_M I_M'), \\
I_Y &= 0.5 (H_L I_L' + H_M I_M'), \\
I_B &= I_S', \\
\end{split}
\label{eq:xyb}
\end{equation}
\noindent
where $H_L \simeq 1$ and $H_M \simeq 1$ are fixed constants~\cite{bb:opsin}.

\subsection{Noise estimation}

We estimate noise during the encoding step of our algorithm. In short, we first
select homogeneous patches from the image, that are the ones that do not contain texture, edges,
or other image details. Then, we design and train the noise model based on the intensity values
of these patches. In the remainder of this section we discuss this process in more detail.

\begin{figure}[t!]
\centering
\includegraphics[width=10cm]{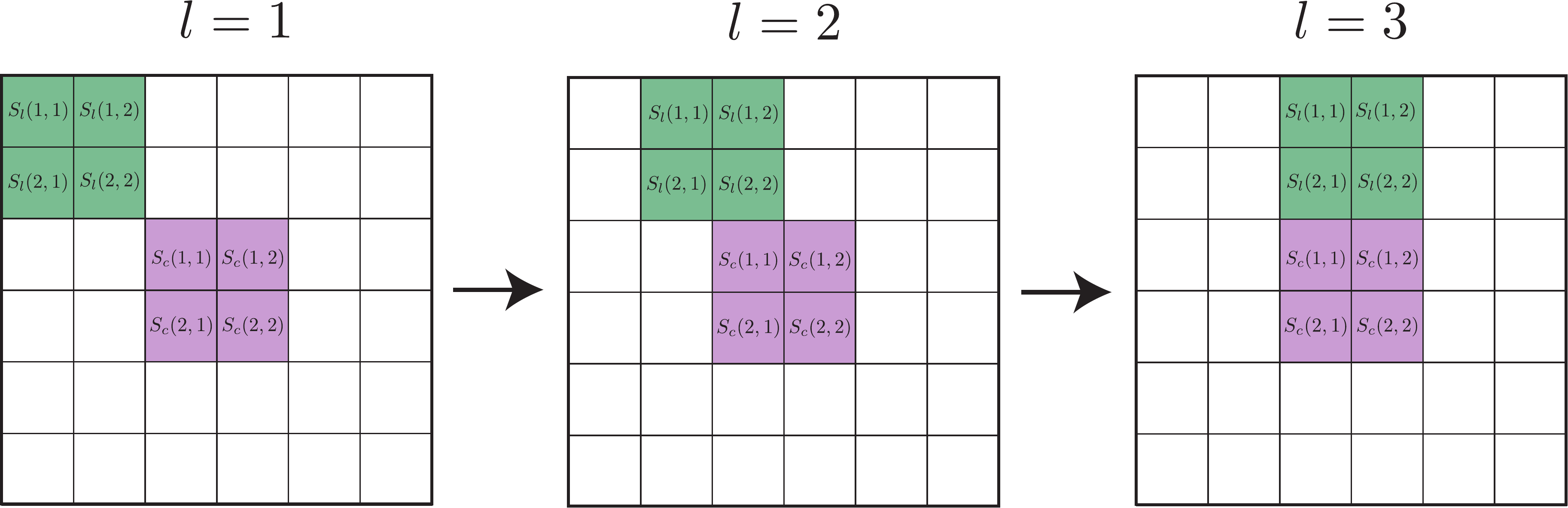}
\caption{Illustration of SAD points evaluation. To calculate $s^{(p)}_\textbf{SAD}(l)$
metric for block $l \in [1,..,K]$ in patch $p$ we calculate the sum of
differences between central block $S_c$ and $S_l$}
\label{fig:sad}
\end{figure}

\paragraph{Homogeneous patches.}
Noise is difficult to separate from image details. Therefore, texture, edges, and
spots inside an image patch may result in imprecision of the noise estimation
algorithm. In order to precisely estimate the level of image noise we first
select ``homogeneous'' patches that are free from the aforementioned image
details. To do so we divide the image patch into a set of blocks $S_l, l \in
[1..K]$, as shown in Fig.~\ref{fig:sad}. Further, following the work~\cite{bb:richardson2004h}, for each patch of the image $p$ we calculate the Sum of Absolute Differences (SAD) similarity measure between center block $S_c$ and each of $S_l, l \in [1..K]$ as follows:
\begin{equation}
s^{(p)}_\textbf{SAD}(l) = \sum_{i=1}^{K_i}\sum_{j=1}^{K_j}|S_c(i,j) - S_l(i,j)|,
\label{eq:SAD}
\end{equation}
where $S_c, S_l$ are the two $[K_i \times K_j]$ image blocks of pixel
intensities, as described in Fig.~\ref{fig:sad}. Inspired by the Rank-Ordered Absolute Differences (ROAD)
metric~\cite{bb:garnett2005universal}, which is shown to be robust to impulse
noise,
we then create a subset $\Omega$ from the $K/2$ elements of  $s^{(p)}_\textbf{SAD}(l),
l \in [1, .., K]$ which have the smallest values.
The resulting \emph{homogeneity} measure $r(p)$ can be computed for each patch
as:
\begin{equation}
r(p) = \frac{2}{K} \sum_{l\in[1 .. K]: s^{(p)}_\textbf{SAD}(l)\in\Omega} s^{(p)}_\textbf{SAD}(l).
\label{eq:SAD_ROAD}
\end{equation}

\begin{figure}[t!]
\begin{tabular}{ccc}
\includegraphics[height=2.2cm,width=3.2cm]{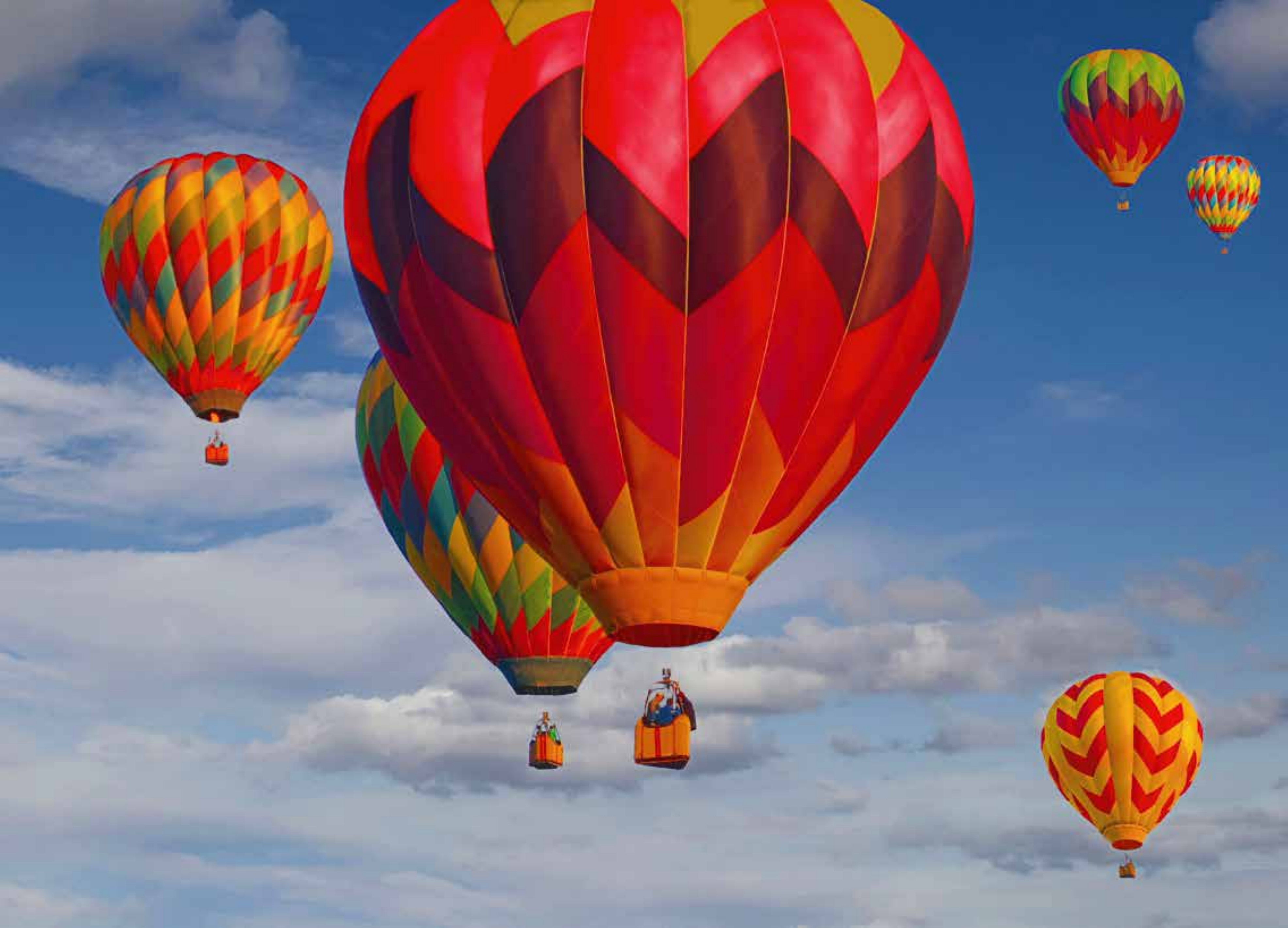}&
\includegraphics[height=2.2cm,width=3.2cm]{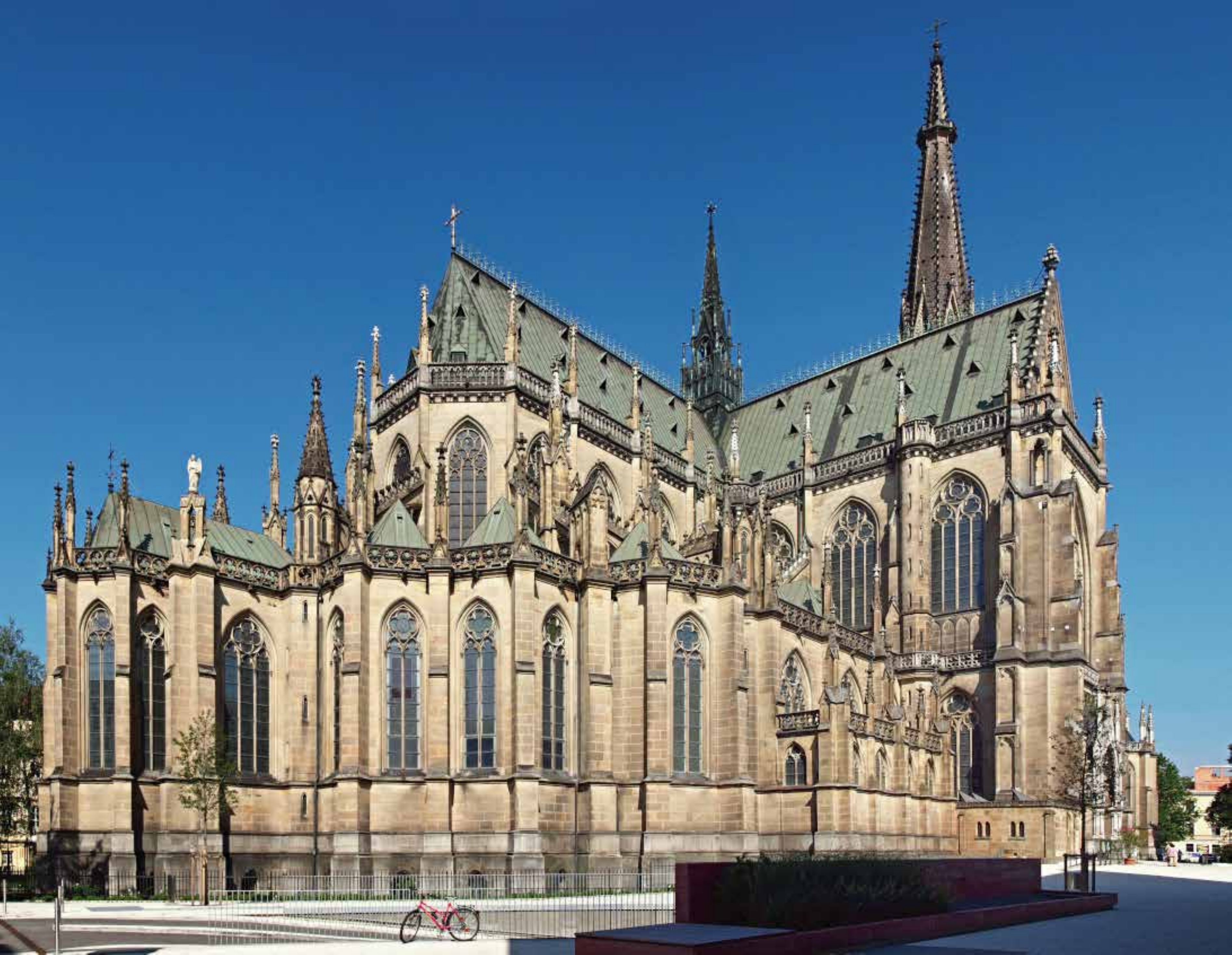}&
\includegraphics[height=2.2cm,width=3.2cm]{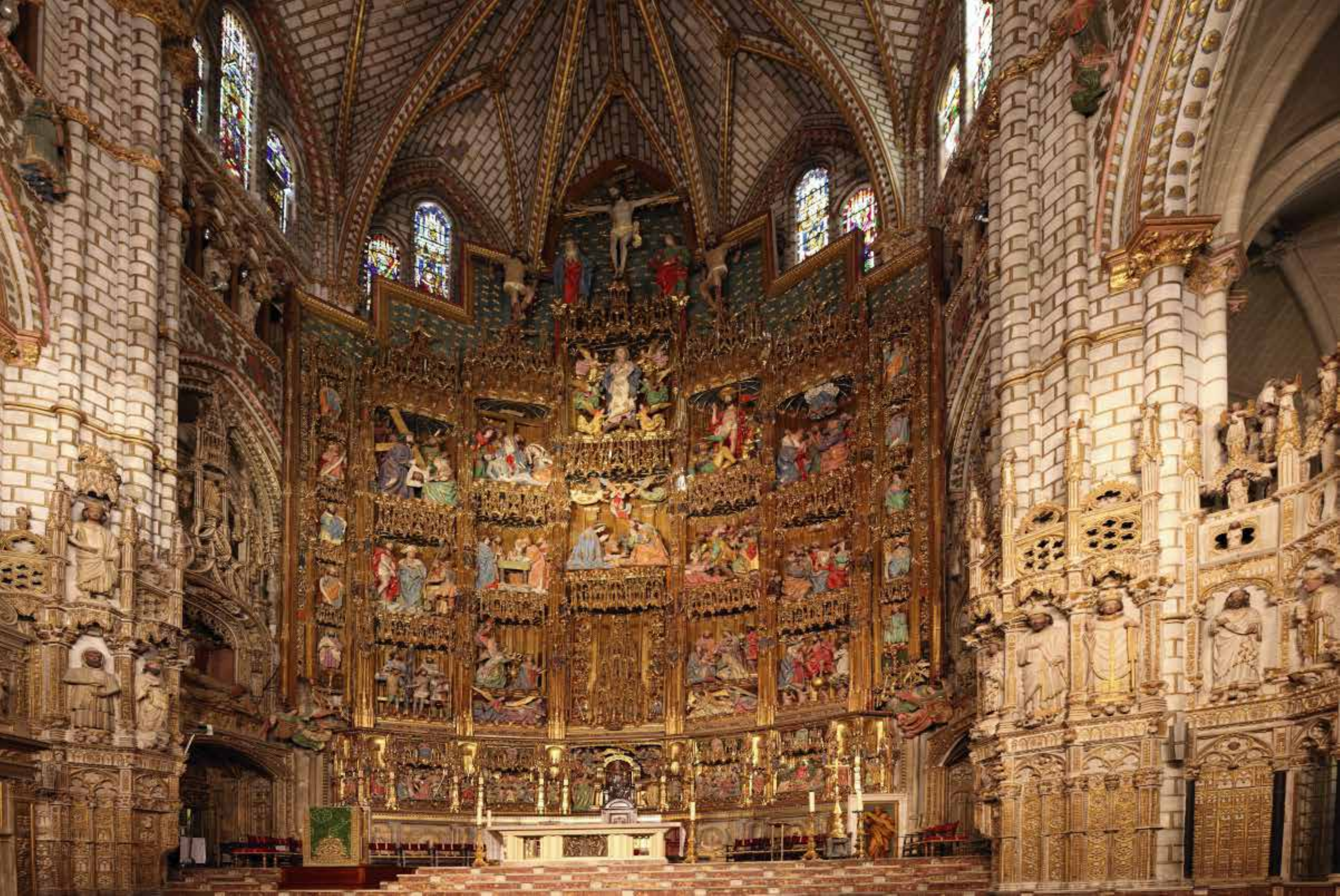}\\
\includegraphics[width=4.1cm]{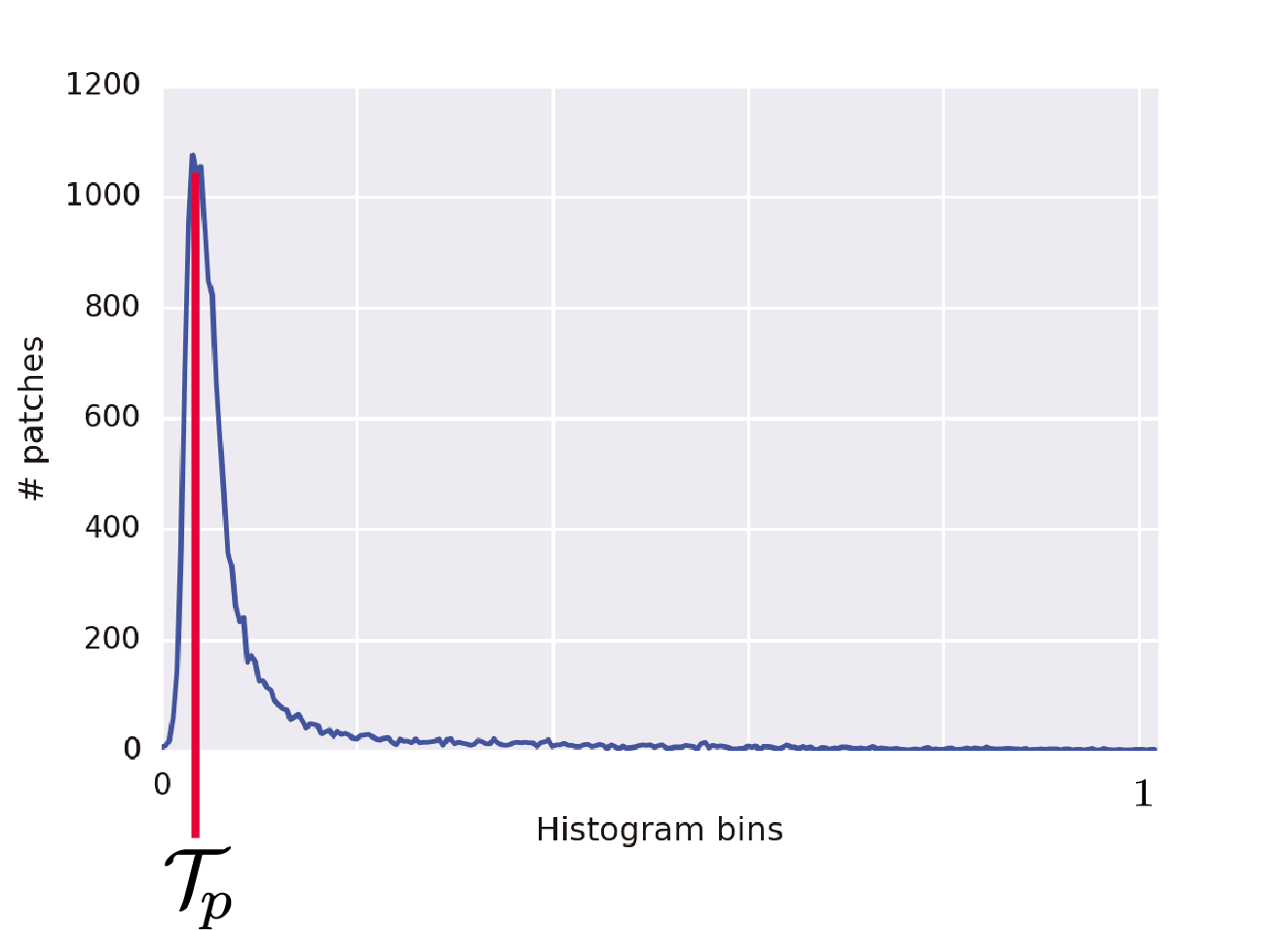}&
\includegraphics[width=4.1cm]{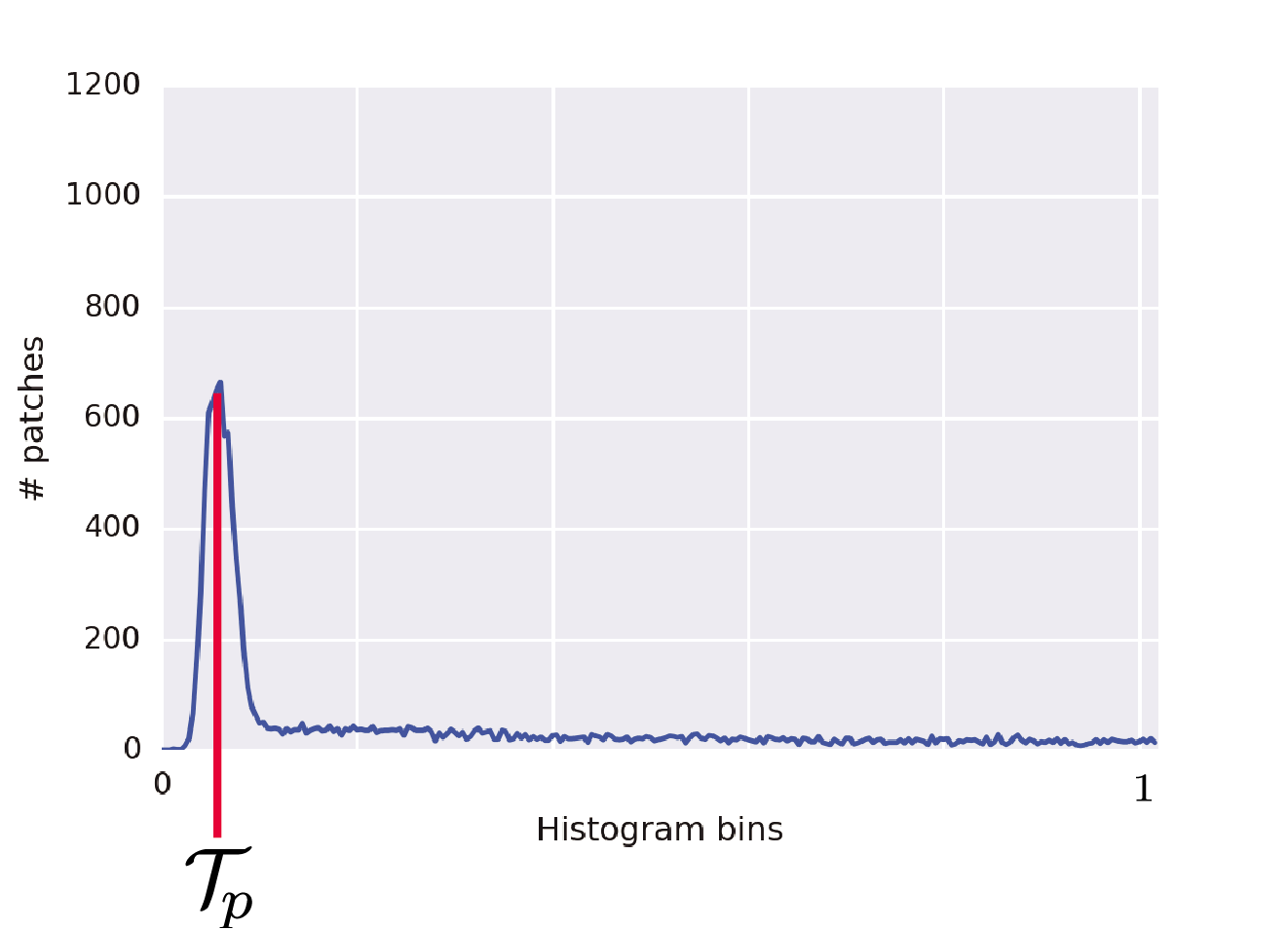}&
\includegraphics[width=4.1cm]{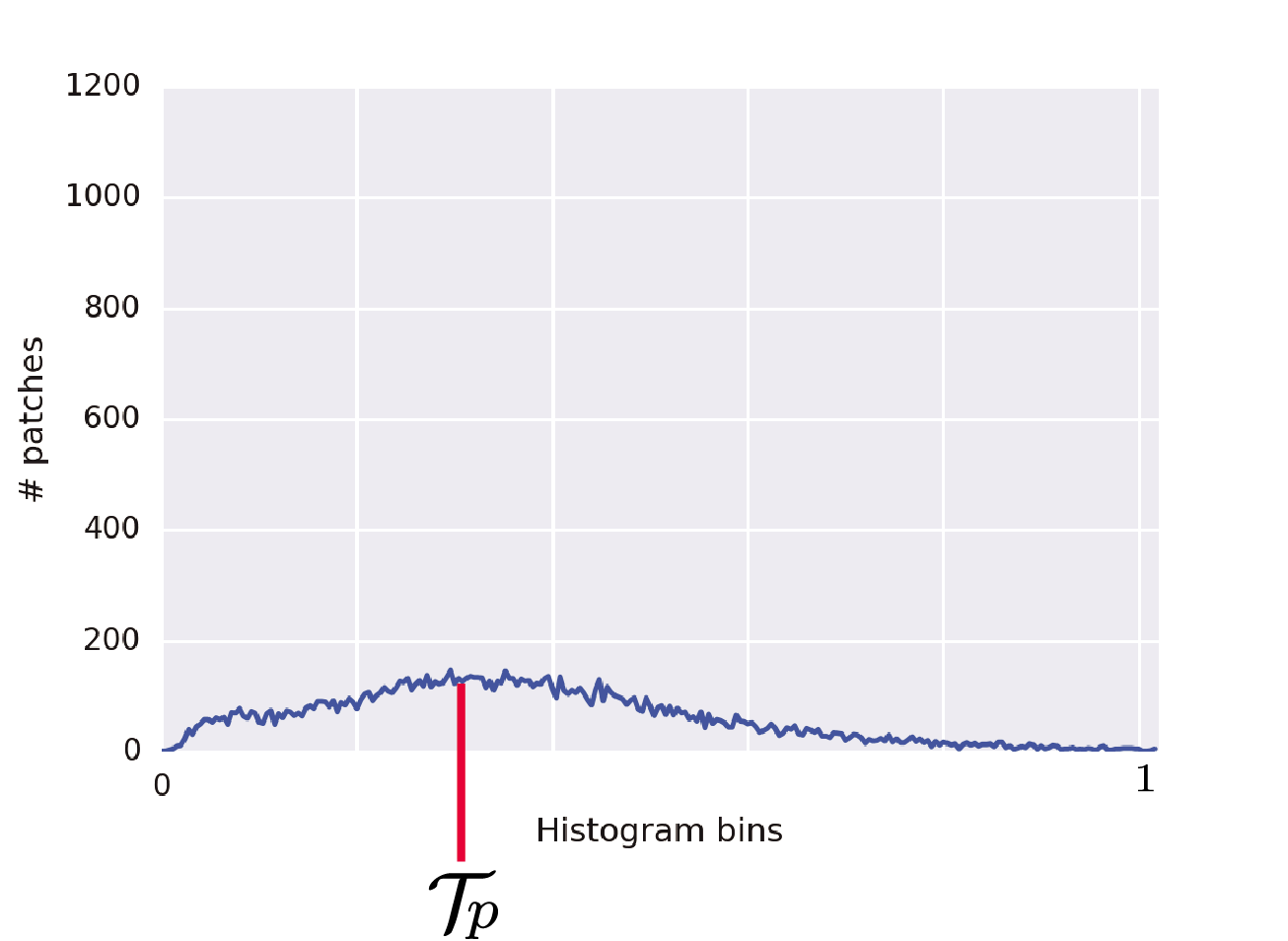} \\
\end{tabular}
\caption{Example of images [\textsc{Top}] and their SAD histograms [\textsc{Bottom}]. Each
histogram has a peak, which correspond to the smoothest part of images if image
contains large enough amount of homogeneous patches. We use this peak value to
choose threshold~$\mathcal{T}$}
\label{fig:histogram}
\end{figure}

Using this metric, we can estimate whether the patch is
homogeneous or not, based on Eq.~(\ref{eq:SAD_ROAD}). This can be done with a
simple threshold $\mathcal{T}$, such that for the patch $p$ to be homogeneous
the following condition need to be fulfilled:
\begin{equation}
r(p) < \mathcal{T} \;.
\end{equation}
\noindent
This approach, however, requires manual selection of the $\mathcal{T}$ value,
which depends on the image properties.
To overcome this limitation we build a histogram of $r(p)$ values from the image
patches. Fig.~\ref{fig:histogram} illustrates sample images with the respective
histograms. As we can see, histograms of images which contain homogeneous areas
have a large peak. Natural images typically have only a single peak, however,
if an image contains multiple we choose the largest one.
This peak corresponds to a set of patches that have low
$r(p)$ scores. Therefore, in order to automatically select a homogeneity threshold
$\mathcal{T}$ for such images we simply need to find the location of this peak
in each of the histograms $\mathcal{T}_p$, which we compute using a robust mode
estimator~\cite{bb:bickel2006fast}.
However, for images that constitute a high level of texture and detail, the peak
of the histogram may correspond to a high $r(p)$ value, which no longer corresponds
to a set of homogeneous patches and therefore may lead to erroneous noise model estimation.
In this work we overcome
this problem by empirically setting the maximum possible threshold
$\mathcal{T}_\text{max}$.
Thus, we compute the homogeneity threshold as follows:
\begin{equation}
\mathcal{T} = \min{(\mathcal{T}_p, \mathcal{T}_\text{max})}
\end{equation}
\noindent

\paragraph{Noise level.} Given a set of homogeneous patches we need to estimate
the noise level of the original image that will later on allow us to re-generate
the same amount of noise to add to the decompressed image. To do so we define
the noise level metric as:
  \begin{equation}
  \hat{n}(p) = \frac{1}{M} ||\mathcal{L} * I(p)||_{1},
  \label{eq:lapl_norm}
  \end{equation}
  where $I(p)$ is the matrix of pixel intensity values of patch $p$,
$\mathcal{L}$ is a Laplacian filter \cite{bb:laplfilt}, $M$ is the total number of pixels in each patch and
$*$ is the convolutional operator. The Laplacian filter $\mathcal{L}$ is a discrete version of the Laplace operator and is defined as follows:
  \begin{equation}
  \mathcal{L} :=
  \begin{bmatrix}
      0 &  1 & 0 \\
      1 & -4 & 1 \\
      0 &  1 & 0 \\
  \end{bmatrix}
  .
  \label{eq:lapl}
  \end{equation}
%  The weight of the central pixel is equal to $w_{0,0} = \sum_{i,j) \neq (0,0), i, j \in\{-1,0,1\}}w_{i,j}$ and
%weights of its neighbours are chosen based on the following kernel:
%  \begin{equation}
%  w_{i,j} = -\exp{\big(\vartheta(1 - \sqrt{i^2 + j^2})\big)}, (i,j)\neq(0,0), i,j\in\{-1,0,1\},
%    \label{eq:lapl_weights}
%    \end{equation}
% which depends on the distance between pixels and relies on $\vartheta$ as a weighting parameter.
Therefore, the intuition behind this noise level metric (Eq.~(\ref{eq:lapl_norm})) resides in the nature of the Laplace
operator, which describes the divergence of the gradient of the image signal.
As a result, $n(p)$ is close to $0$ for smooth patches and large for the noisy
ones. Further, due to the fact that
  $\sum_{i,j} \mathcal{L}_{i,j} = 0$ the metric does not depend on the
pixel's intensity values, which essentially means that the same additive noise
gives the same score for various intensity values. Thus, the metric provides an
unbiased estimate of the noise level for different homogeneous patches.

\paragraph{Noise model.}
In both camera sensors and human eyes, noise resembles a Poisson
distribution~\cite{bb:verma2013comparative}. Therefore, in principle, image
noise can be modeled using the intensity dependent model of~\cite{bb:verma2013comparative}.
However, it does not take into account various camera post-processing steps,
which include gamma correction and
de-mosaicking. Further, as mentioned earlier, image noise typically depends on
material properties of objects that are present in the scene and, therefore,
varies across different parts of the image. To alleviate these issues we suggest to
learn an intensity-dependent model for each image as follows:
\begin{equation}
n(I_L') = \alpha {(I_L')}^\gamma + \beta,
\label{eq:noise_model}
\end{equation}
where $\alpha$, $\beta$ and $\gamma$ are the parameters of the noise model. These
parameters are trained independently for each image during the encoding step by
minimizing the following objective function:
\begin{equation}
\mathcal{F} = \sum_{p\in \mathcal{H}}\bigg{(}\hat{n}(p) - n\big{(}\overline{I}_L'(p)\big{)}\bigg{)}^2 + \xi \alpha \gamma,
\label{eq:loss}
\end{equation}
where $\mathcal{H}$ is a set of homogeneous patches,
$\overline{I}_L'(p)$ is the mean intensity of the patch $p$ for the channel
$L$ from Eq.~(\ref{eq:lms_xyb}), $\hat{n}(p)$ is the noise level,  calculated by
Eq.~(\ref{eq:lapl_norm}), $ n\big{(}\overline{I}_L'(p)\big{)}$ the noise model,
defined in Eq.~(\ref{eq:noise_model}) and $0 \leq \xi < 1$ is the empirically chosen weight of the
regularization parameter $\alpha \gamma$. We minimize this function using the
scaled conjugate gradient method \cite{bb:moller1993scaled}. It is worth noting
that we chose the power function as our noise generation model because we work
with gamma corrected signal $\overline{I}_L'(p)$ in g-LMS space. In practice such
a model is very compact and requires only a few bytes of additional memory
regardless of image size, as it effectively needs storing just three
(quantized) floating-point values: $\alpha, \beta$, and $\gamma$.

\subsection{Noise re-generation}

Finally, during the decoding step, we re-generate noise using estimated
parameters $\alpha, \gamma, \beta$ from Eq.~(\ref{eq:noise_model}) and intensity
values of the image pixels. Briefly, in order to achieve this, we first estimate
the expected noise level for each pixel and then generate random noise, which
satisfies this value. We now discuss this process in more detail.

\paragraph{Noise level estimation.}
As discussed in Section \ref{s:alg_xyb} the XYB color space has a direct
relationship with the g-LMS space. This allows us to add different amount of
noise to different wavelengths of the input signal. Due to the nature of the
human eye, the signals with long and medium wavelengths matter the most,
therefore here we consider only channels $I_{L}'$ and
$I_{M}'$ of the g-LMS space, introduced in Eq.~(\ref{eq:lms_xyb}). Based on our
noise model, we then estimate the appropriate noise level for each of these
channels as:
\begin{equation}
\begin{split}
n_M = n(I_M'), \\
n_L = n(I_L'),
\label{eq:noise}
\end{split}
\end{equation}
where $n_M, n_L$ depict the noise levels for each pixel of the g-LMS channels
corresponding to medium and long wavelengths respectively.

\begin{figure}[t!]
\centering
\includegraphics[height=1.4cm]{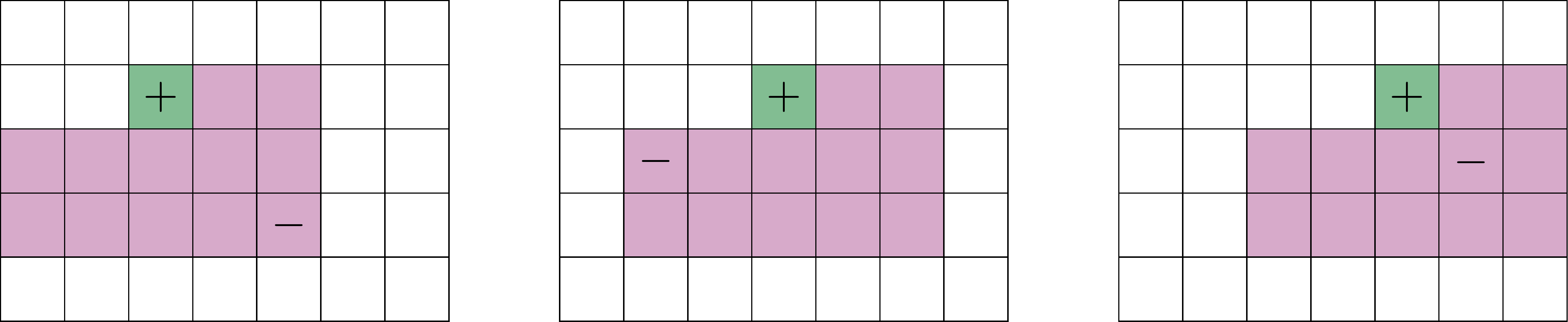}
\caption{Matrix generation process. We first populate each of the elements of
matrices $R_j, j \in [L,M,c]$ with uniform noise. Further we process them
pixel by pixel (shown as ``+'') by subtracting the value ``--'' of a randomly chosen pixel from the
neighborhood}
\label{fig:laplacian_like_subtr}
\end{figure}

\paragraph{Additive random noise generation.}
Now that the noise level for each pixel is estimated, our final step is to
generate the appropriate amount of random noise and add it to the decompressed
image. According to Eq. (\ref{eq:xyb}) both X and Y channels of XYB
color space depend on long and medium wavelength channels of g-LMS space.
Therefore it is natural to assume that the additive noise should contain two
parts, one of which is shared across X and Y, while the other is not. To model
this behavior we generate three random matrices $R_L, R_M, R_c \in
\mathbb{R}^{N_w \times N_h}$ for each decompressed
image with size $N_w \times N_h$. These matrices must contain a high-frequency signal, therefore, to
generate them we apply a high pass filter to random matrices, elements of which
are uniformly distributed in $[0,1]$. To make the computation fast, which is
crucial for a compression algorithm,  we use the following trick. For each element
of a generated matrix $R_{j}, j \in [L,M,c]$  we subtract a random element from
its neighborhood as depicted by Fig.~\ref{fig:laplacian_like_subtr}. Then, the
matrices are normalized to have the following property: $\frac{1}{N_w N_h}||\mathcal{L} * R_{j}||_1
= 1, j \in [L,M,c]$, where $\mathcal{L}$ is the Laplacian matrix, introduced in
Eqs. (\ref{eq:lapl_norm}) and (\ref{eq:lapl}). In such formulation, matrices
$R_L$ and $R_M$ account for independent noise appearing in the long and medium
wavelength channels of g-LMS space, while $R_c$ is used to model correlated
random noise.

Further, we augment each channel of every pixel of the decompressed image in the
following way:
\begin{equation}
\begin{split}
I_X = I_X + \psi (n_L R_L - n_M R_M) + (1-\psi) R_c (n_L - n_M), \\
I_Y = I_Y + \psi (n_L R_L + n_M R_M) + (1-\psi) R_c (n_L + n_M), \\
I_B = I_B + H_B \psi (n_L R_L + n_M R_M) + (1-\psi) R_c (n_L + n_M), \\
\label{eq:L}
\end{split}
\end{equation}
where $n_L$, $n_M$ are the noise levels, introduced in Eq.~(\ref{eq:noise}) and
$\psi : 0 \leq \psi \leq 1$ is the regularization parameter, which allows
balancing the correlation between noise, generated for the X and Y channels of
the XYB color space. In practice, parameter $\psi$ has direct influence on the
`colorfulness' of the generated noise, with the degenerate case of $\psi = 0$
that corresponds to the generation of completely gray-scale noise.

\section{Experiments}
\label{s:exp}

In this section we describe our experiments. First, we describe the settings
and the parameters of the algorithm. Then, we show visual results of our approach.
Finally, we present the results of the user study that aims at quantifying the improvement
in quality of visual perception of images processed with our noise re-generation
algorithm. The code for the proposed approach is available under the following link: \url{https://github.com/google/pik}.

\subsection{Settings}

We run our experiments with the recently proposed lossy compression algorithm
PIK~\cite{bb:pik}, which is designed to replace JPEG~\cite{bb:wallace1992jpeg}
with about one-third the data size at similar perceptual quality. Our goal is to improve
visual perception of strongly compressed images as for these conditions compression
algorithms often remove important small image details.
Therefore, we re-generate noise in the PIK output with Butteraugli psychovisual target
distance~\cite{bb:opsin} equal to 3.0. This setting results in highly compressed images, which is
achieved at the cost of removal of some image details and introduction of
ringing and blocking artifacts. To reduce the influence of these effects, in our
experiments, we apply a simple deblocking filter similar to
\cite{bb:norkin2012hevc} before adding noise.

In order to evaluate our method, we have created the dataset of images with different
noise conditions. In particular, we have been using images from~\cite{bb:Wikimedia}~\footnote{We use the images that can be found under the following link: \\ \scriptsize{https://github.com/WyohKnott/image-formats-comparison/blob/gh-pages/cite\_images.txt}}. Furthermore, we describe the
parameters of our method for noise estimation and re-generation below.

\paragraph{Noise estimation parameters.}
We set the size of image patches, which are used to estimate noise, to $8
\times 8$ which is the same as PIK's block size. Further, to select the homogeneous
patches we evaluate the SAD metric with sub-patches of size $K_i \times K_j = 3
\times 4$ (see Eq.~(\ref{eq:SAD})). To build the SAD histogram we use $\mathcal{T}_\text{max} =
0.2$ and $255$ bins.

\begin{figure}[t!]
\begin{tabular}{cccc}
\includegraphics[width=2.6cm, height=1.6cm]{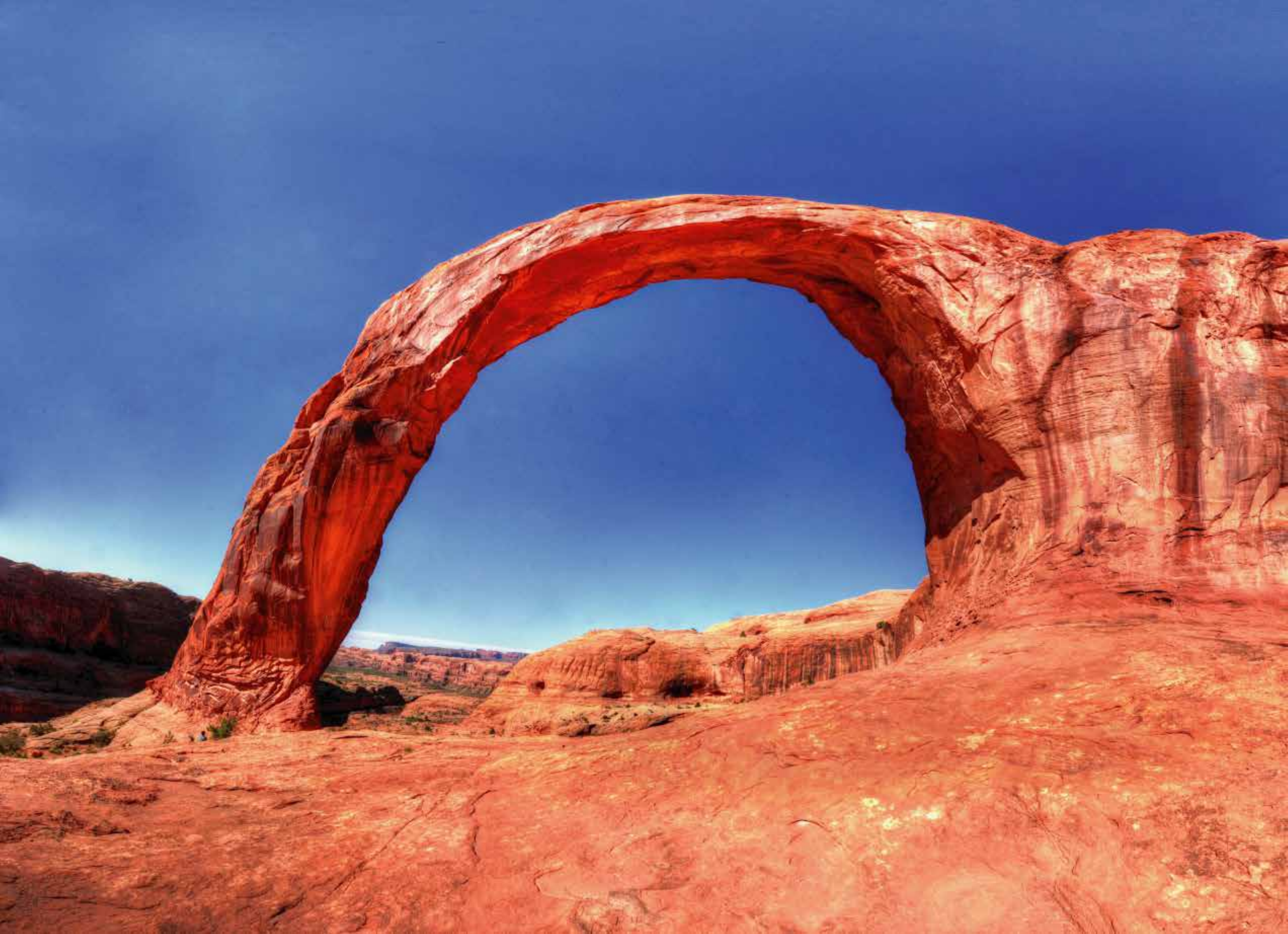}&
\includegraphics[width=2.6cm, height=1.6cm]{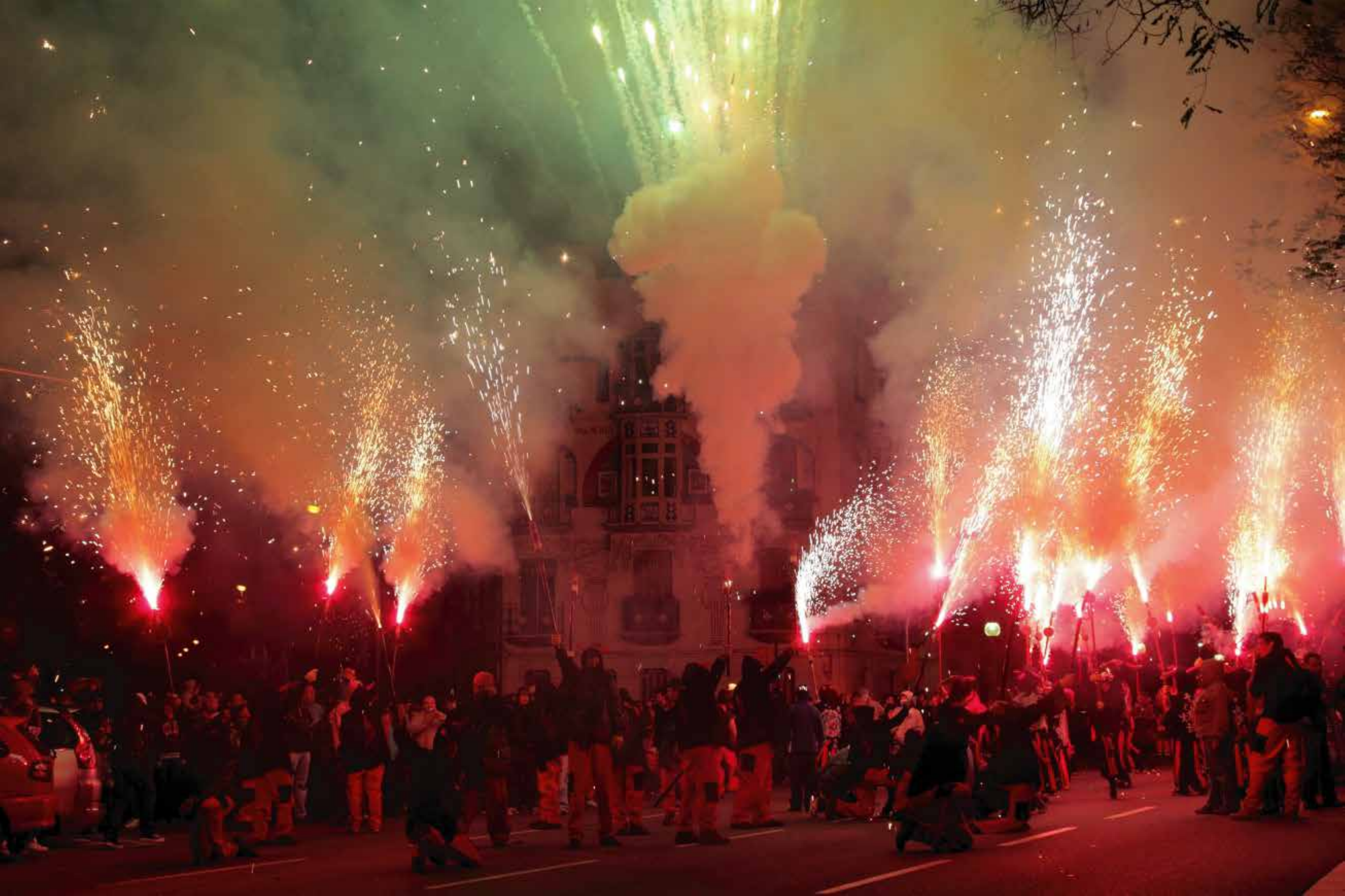}&
\includegraphics[width=2.6cm, height=1.6cm]{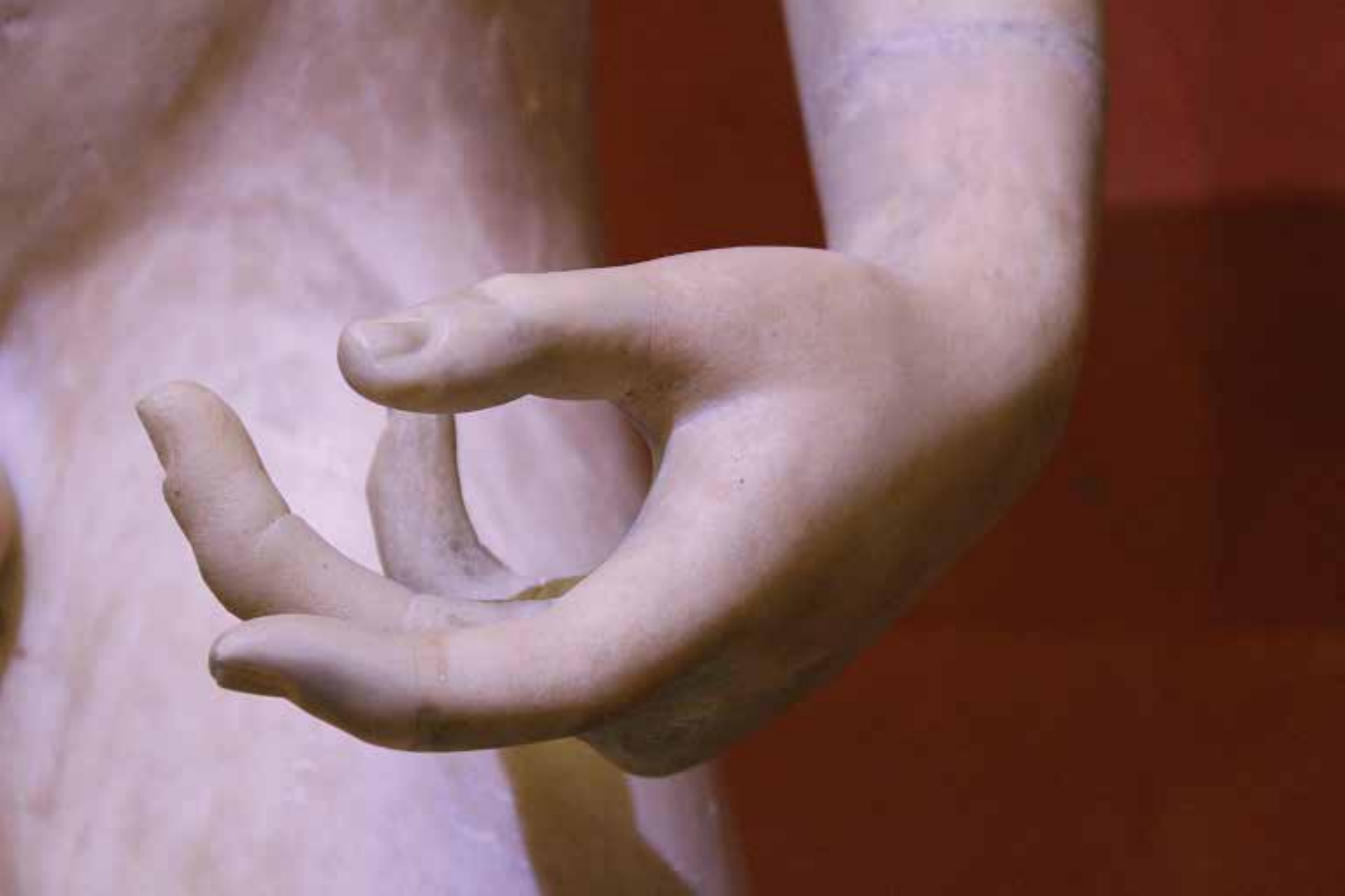}&
\includegraphics[width=2.6cm, height=1.6cm]{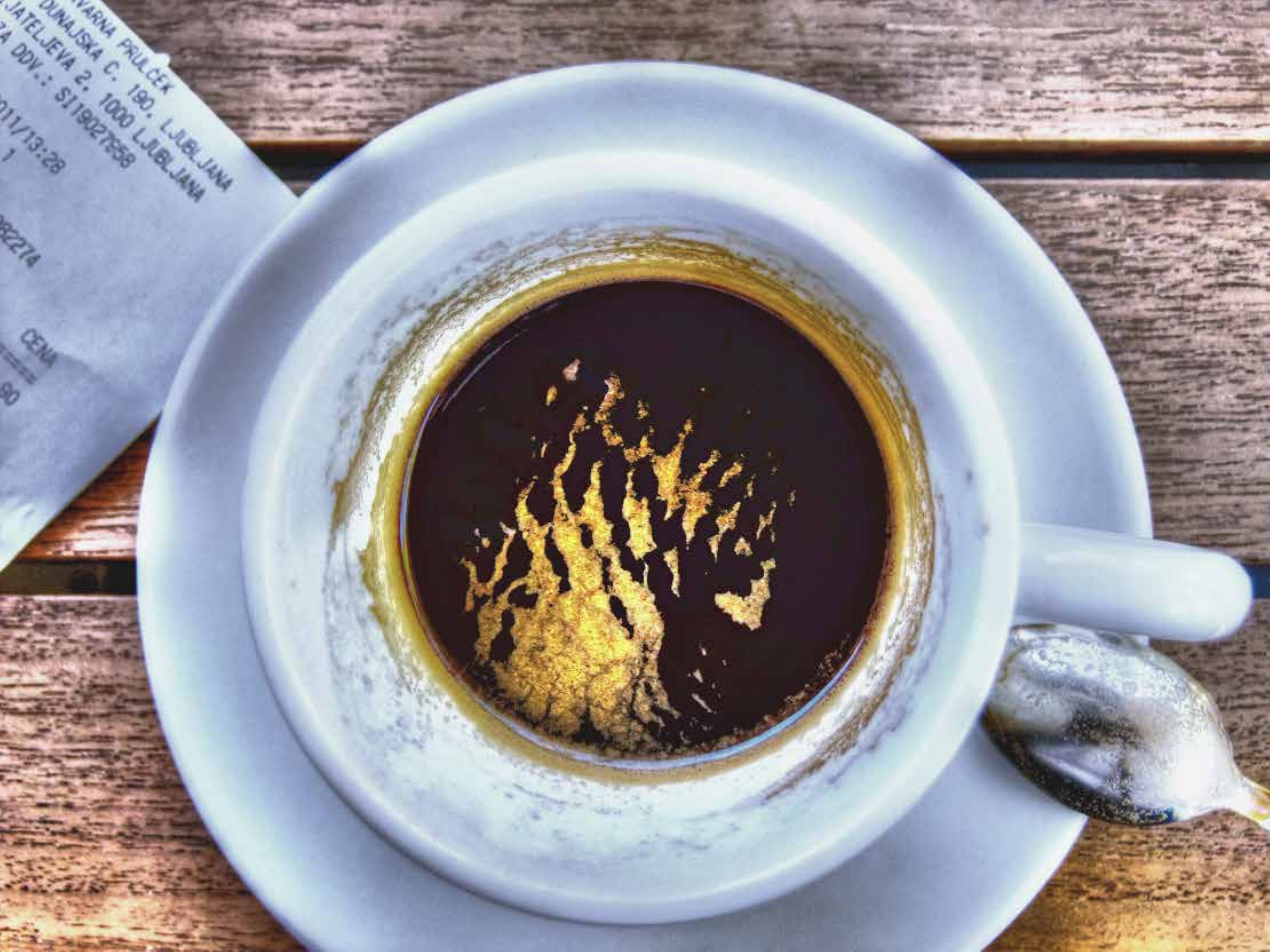}\\
\includegraphics[width=3cm]{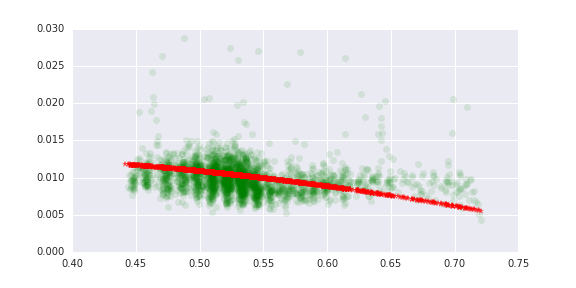}&
\includegraphics[width=3cm]{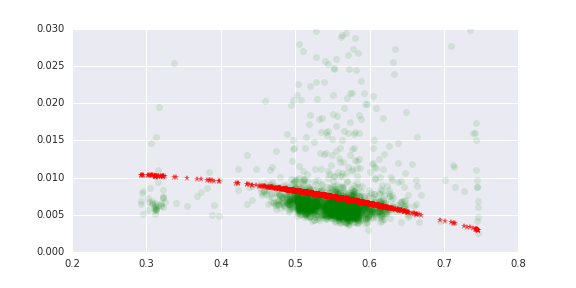}&
\includegraphics[width=3cm]{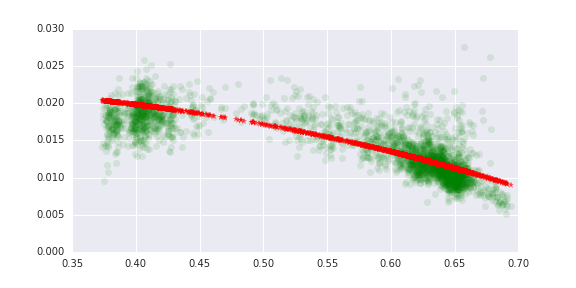}&
\includegraphics[width=3cm]{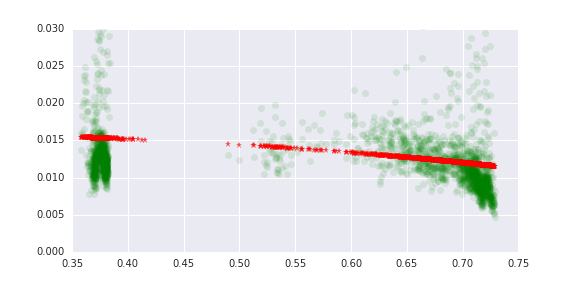}\\
(a)&(b)&(c)&(d)
\end{tabular}
\caption{Examples of images [\textsc{Top}] and their respective noise models [\textsc{Bottom}].
Green dots correspond to the noise estimated from the image, while the red
stars illustrate the model estimated from these points. (best seen in color)}
\label{fig:noise_estimation}
\end{figure}

Sample images with the respective noise modes are illustrated by Fig.~\ref{fig:noise_estimation}.
It is worth noting that in the linear color space the amount of noise grows with signal intensity.
However, due to the fact that modern cameras apply gamma correction, the relative noise level in digital images is
considerably stronger in the dark areas, rather than in the bright ones. A similar effect is happening in a human eye,
which makes people perceive noise differently depending on the areas of different brightness.
To model this effect, we have introduced a regularization parameter $\xi = 5 \times 10^{-5}$
to the Eq.~(\ref{eq:loss}). This regularization allows optimization to fit the
model to the data, giving priority to decreasing functions.

\paragraph{Noise re-generation parameters.}
The generation part of the proposed algorithm relies on a color adjustment
parameter, which we set to $\psi = 0.1$. Lower value of $\psi$ gives less
colored noise and, according to our preliminary experiments, this value results in images that are the most pleasant for users.

\subsection{Visual results}

\begin{figure}
\begin{tabular}{cccc}
\raisebox{2cm}{a)} &
\includegraphics[width=4cm, height=4cm]{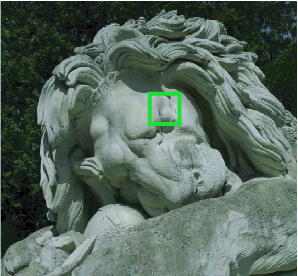}&
\includegraphics[height=4cm, width=4cm]{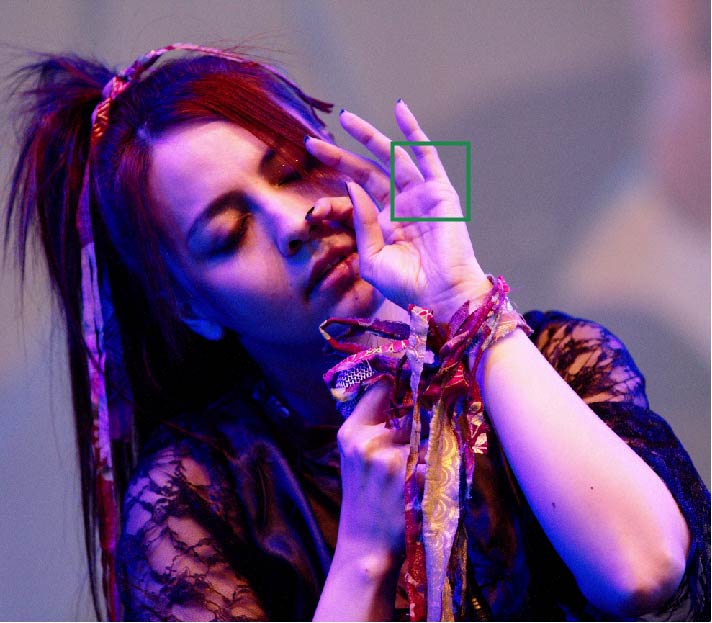} &
\includegraphics[width=4cm, height=4cm]{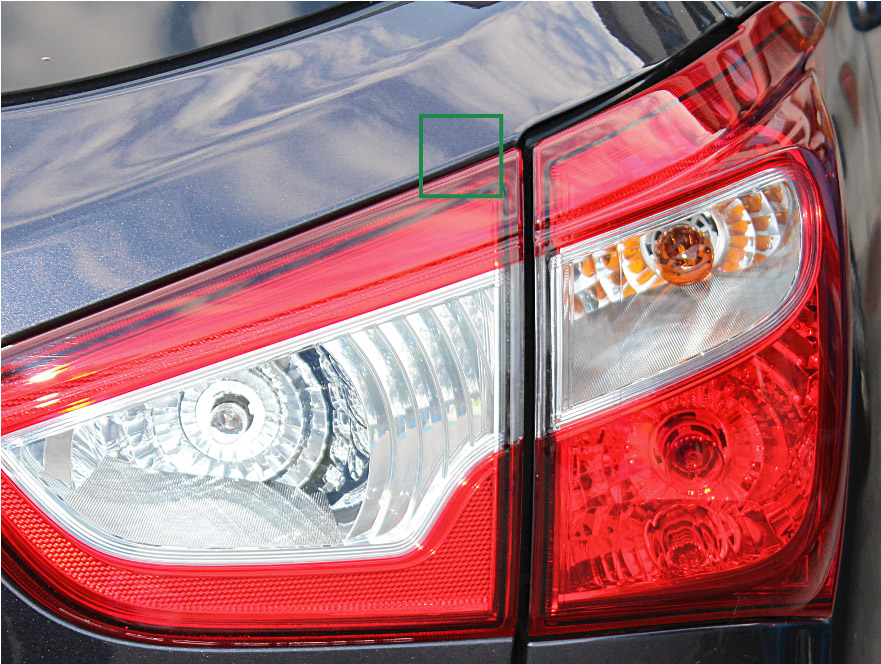}\\

\raisebox{2cm}{b)} &
\includegraphics[width=4cm]{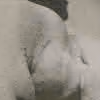}&
\includegraphics[height=4cm]{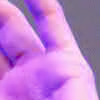} &
\includegraphics[width=4cm]{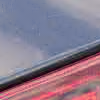}\\

\raisebox{2cm}{c)} &
\includegraphics[width=4cm]{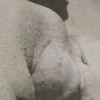}&
\includegraphics[height=4cm]{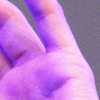} &
\includegraphics[width=4cm]{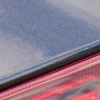} \\

\raisebox{2cm}{d)} &
\includegraphics[width=4cm]{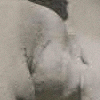}&
\includegraphics[height=4cm]{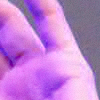} &
\includegraphics[width=4cm]{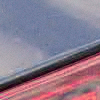}\\
\end{tabular}
\caption{Evaluation of our algorithm with respect to PIK. The first row (a)
illustrates sample images to which our method is applied. Further for each image,
we show three zoomed-in patches (b-d), where patch (b) corresponds
to the result of the PIK algorithm, (c) is extracted from the original image and
(d) depicts the result of our noise generation technique applied to the image
decompressed by PIK algorithm. (best seen in color)}
\label{fig:visual}
\end{figure}

In this section we show visual results of our algorithm. To better illustrate the
advantage of our technique, we select images with some
high-frequency signal, as it is altered the most by the compression algorithm.
Fig.~\ref{fig:visual} illustrates the performance of our approach.
In particular the first row illustrates the original high resolution image.
Then the middle two rows show the image patches cropped from the decompressed and
original images respectively. The last row illustrates the patch from the image
produced by our algorithm, which appears to be much closer to the original image
(third row) than the decompressed one (second row).

\subsection{User experiment}

In order to evaluate quality of the visual perception of the images
generated by our method, we run three experiments. The first two aim at determining if
users prefer images with noise compared to their smooth versions (provided
by the PIK compression algorithm). For these experiments, we select images and
generate noise for them with noise levels chosen by an expert. Finally, in
our last experiment we evaluate how close the noise level estimated by our
system is to the one preferred by users.
%We conducted our user study on the colleagues
%that were not involved in this research project.

\begin{figure}
\centering
\begin{tabular}{cc}
\includegraphics[height=3.4cm]{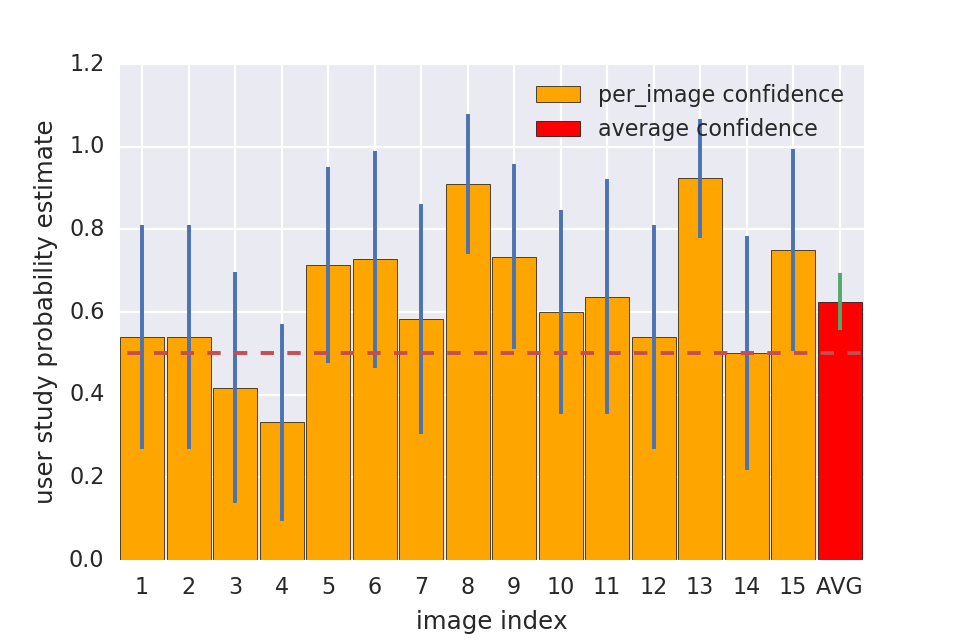} &
\includegraphics[height=3.4cm]{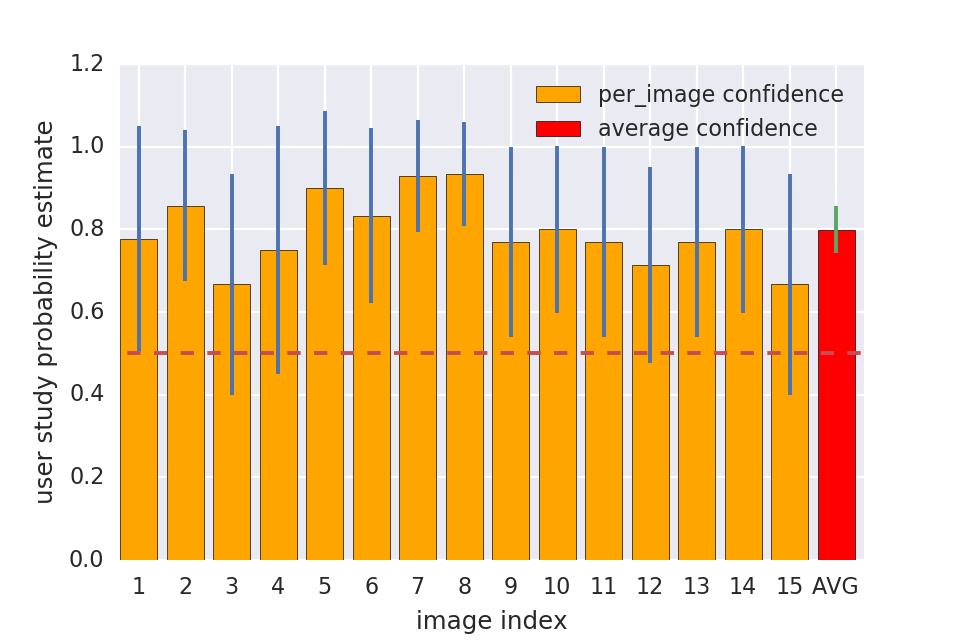} \\
(a) Perceived quality & (b) Authenticity
\end{tabular}
\caption{Probability estimation of choosing the image produced by our noise
generation algorithm with 95\% confidence level for the first (a) and second
(b) experiment. Last bar in both figures corresponds to the average probability
estimate across all images and users}
\label{fig:perc_qual}
\vspace{-0.35cm}
\end{figure}

\paragraph{Perceived quality of noise generation.}
For this experiment we select a dataset of $15$ pairs of images and show them to
$15$ people. Each pair contains an image processed by the PIK algorithm and another one with
additive image noise, generated by our approach. During the experiment, users are
asked to choose the image that they perceive to be of higher quality, without
knowing which one of the two corresponds to our method. This allows evaluating
the performance of our approach according to user preferences. The result of this
experiment is depicted by Fig.~\ref{fig:perc_qual}(a), which for each pair of
images shows the probability with the $95\%$ confidence interval of users preferring image
with noise over the one without it. The average probability is
then depicted by the final column in Fig.~\ref{fig:perc_qual}(a). As we can see,
users often prefer noisy images as compared to smooth ones.

\paragraph{Authenticity of noise generation.}
For this experiment we slightly change the conditions of the aforementioned
experiment. Here, instead of pairs of images we have triplets that contain
the original image, the one processed by the PIK algorithm and another one
with our generated noise. These triplets are then shown to $15$ users, who
need to choose the most authentic copy of the original image between the image
processed by PIK and the one generated by our method, without knowing which one
is which. The result of this experiment is summarized by Fig.\ref{fig:perc_qual}(b),
which for each triplet illustrates the probability with its confidence interval of
how likely the image with the noise will be selected. The last column then illustrates
the average probability across all triplets in the dataset. As we can see, users
typically prefer images with the noise, as according to their perception they look
closer to the original ones, over images produced by the PIK
algorithm with no additive noise.

These two experiments show that our method generates noise that in most of the
cases is pleasant for the users and makes decompressed images look more natural.

\begin{figure}
\centering
\begin{tabular}{cc}
\includegraphics[height=3.4cm]{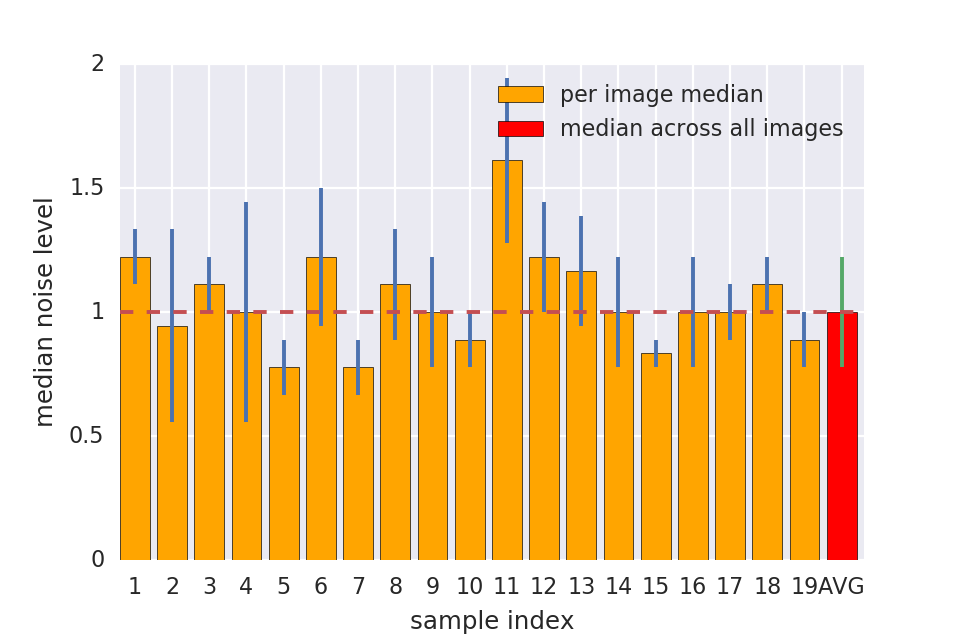} &
\includegraphics[height=3.4cm]{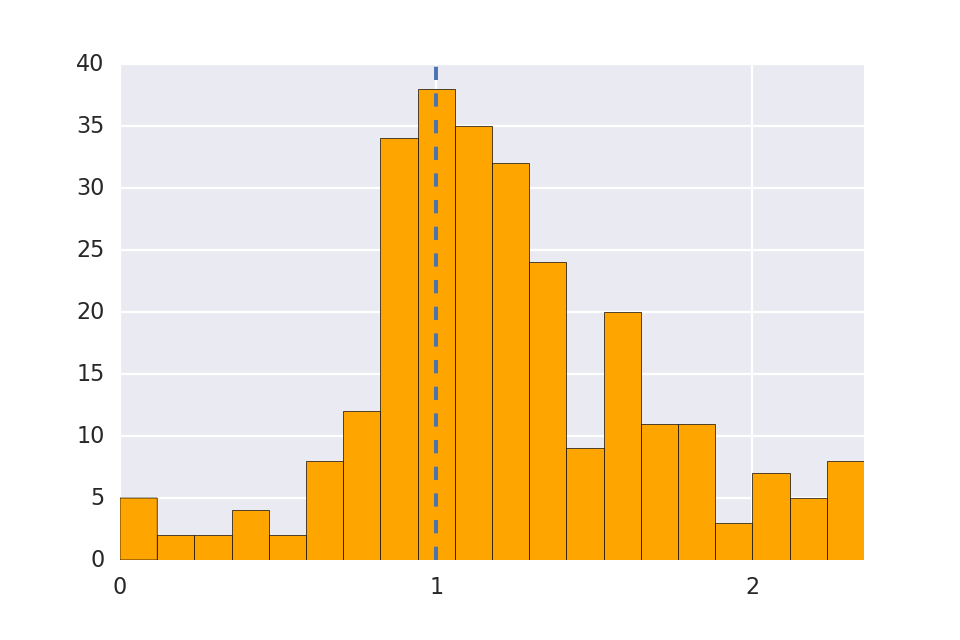}
\end{tabular}
\vspace{-0.15cm}
\caption{ [\textsc{Left}] Median noise level with respective median
absolute difference, which is chosen by 15 users. The last bar corresponds
for the median across all images and users. [\textsc{Right}] Distribution
of preferred noise levels across all images and users}
\label{fig:noise_level_exp}
\vspace{-0.2cm}
\end{figure}

\paragraph{Estimation of noise and noise re-generation.}
For our last experiment we developed a system that allows users to adjust
the level of generated noise that is added to the decompressed image.
Specifically, we allow users to select one of $20$ noise levels to
find the one that makes the decompressed image look as close as possible
to the original image (before compression). These $20$ levels come from the
multiplication of the noise level estimated by our method with the coefficients
starting from $0$ with the step $0.125$.
For this experiments we have selected $19$ examples from our dataset and show
them to $15$ users. The users are then asked to adjust the smoothness
over the graininess of the compressed image to match their viewing
experience with the original one. The results of this experiment are
summarized in Fig.~\ref{fig:noise_level_exp}, which shows the median
and median absolute difference of the noise levels selected by the users
for each of the examples in the dataset. The last column illustrates the
median noise level that was selected across all people and images in the
dataset. As we can see the average noise level is very close to $1$,
which means that our noise generation system allows to estimate the proper
noise level in most of the cases.

To further analyze the performance of our system, we build a histogram,
which illustrates the  distributions of the votes for different noise level
over all users and images. We can see that the majority of the votes are
concentrated near the $1$ value. Further, only very few votes show
preference of noiseless images, which means that users typically prefer
the images with noise generated by our method, as opposed to the smoother
images, produced by the PIK algorithm. This hypothesis is also supported
by the previous experiments.

We further investigated the images, for which users generally prefer having
higher amount of noise. It turned out that these were the highly textured
images, which means that having a model that is solely inspired by the camera
sensor is not enough and more complex models that also take into account image
texture can be used. Application of such models, however, may result in a
significant increase of processing time and memory required for
storing the parameters of the these models, which may become a severe
limitation for compression algorithms. There are also a number of images
in our dataset where the users generally prefer having smaller noise
level than the one suggested by the model. These images typically contain
very shiny surface areas, where the users expect not to see any noise at all.
We would like to address these issues and improve our technique to tackle such
cases in future work.

In summary, our experiments show that users typically prefer images with noise
with respect to the ones that are processed by the PIK algorithm and do not have
any noise at all. Furthermore, our noise estimation model on average selects
a level of noise that is perceived favorably by most users.

\section{Conclusion}
\label{s:concl}

In this paper we proposed a novel noise re-generation method for image compression algorithms,
which estimates the noise model parameters from the input image at the encoding step
and re-generates noise at the decompression step. Our model is physically and
biologically inspired. We further introduced a fast noise generation technique based on a Laplacian
filter which is suitable for fast decompression.
As illustrated by the user study, our method is able to generate the appropriate
level of additive noise that improves the perceived quality of the image.
Our implementation of the proposed algorithm is open-sourced and publicly available.

\clearpage

\bibliographystyle{splncs}
\bibliography{egbib_noise}

\end{document}